\PassOptionsToPackage{prologue,dvipsnames}{xcolor}
\documentclass[sigconf, screen, nonacm]{acmart}

\AtBeginDocument{%
  }

\usepackage{multirow}
\usepackage{arydshln}

\usepackage{amssymb}
\usepackage{graphicx}  
\usepackage[table,xcdraw]{xcolor}  
\usepackage{booktabs}  
\usepackage{amsmath}  
\usepackage{threeparttable}
\usepackage{tcolorbox}
\definecolor{lightcyan}{rgb}{0.48, 1.0, 1.0}
\colorlet{mythmback}{lightcyan!40!white}
\newtcolorbox{boxEnv}{
colback=mythmback,coltitle=blue,colframe=mythmback,
center,
width=0.85\linewidth,
boxrule=0.5pt,
left=10pt,right=0pt,
top=2pt,bottom=2pt,
before skip=10pt, after skip=10pt, 
}

\newcommand{\extractive}[1]{\colorbox[HTML]{CFE2F3}{#1}}
\newcommand{\yn}[1]{\colorbox[HTML]{D9EAD3}{#1}}
\newcommand{\abs}[1]{\colorbox[HTML]{E1D5E7}{#1}}
\usepackage[normalem]{ulem}
\useunder{\uline}{\ul}{}

\begin{document}
\begin{sloppypar}
\title{CAPTURE: A Benchmark and Evaluation for LVLMs in CAPTCHA Resolving}

\author{Jianyi Zhang*, and Ziyin Zhou}
\email{zjy@besti.edu.cn}
\affiliation{%
  \institution{Beijing Electric Science and Technology Institute}
  \city{Beijing}
  \state{Beijing}
  \country{China}
}

\author{Xu Ji, Shizhao Liu, and Zhangchi Zhao}
\affiliation{%
  \institution{Beijing Electric Science and Technology Institute}
  \city{Beijing}
  \state{Beijing}
  \country{China}
}








\begin{abstract}


Benefiting from strong and efficient multi-modal alignment strategies, Large Visual Language Models (LVLMs) are able to simulate human visual and reasoning capabilities, such as solving CAPTCHAs. However, existing benchmarks based on visual CAPTCHAs still face limitations. Previous studies, when designing benchmarks and datasets, customized them according to their research objectives. Consequently, these benchmarks cannot comprehensively cover all CAPTCHA types. Notably, there is a dearth of dedicated benchmarks for LVLMs. To address this problem,  we introduce a novel CAPTCHA benchmark for the first time, named \textbf{\textit{CAPTURE}} (\textbf{CAP}TCHA for \textbf{T}esting \textbf{U}nder \textbf{R}eal-world \textbf{E}xperiments), specifically for LVLMs. Our benchmark encompasses 4 main CAPTCHA types and 25 sub-types from 31 vendors. The diversity enables a multi-dimensional and thorough evaluation of LVLM performance. \textit{CAPTURE} features extensive class variety, large-scale data, and unique LVLM-tailored labels, filling the gaps in previous research in terms of data comprehensiveness and labeling pertinence. When evaluated by this benchmark, current LVLMs demonstrate poor performance in solving CAPTCHAs. 

To enhance LVLMs' proficiency in solving CAPTCHAs, we present a two-stage framework, CRRD (\textbf{C}ropping, \textbf{R}e-\textbf{R}eading, and \textbf{D}escribing). Drawing inspiration from human problem-solving behavior for complex tasks, we first crop CAPTCHA images into instruction text and image, to help LVLM focus on crucial information and eliminate distractions. Then we use the re-reading method to enhance LVLM's reasoning ability, enabling it to better identify key elements in CAPTCHA image. The framework improves LVLM performance across all CAPTCHA tasks, with up to a 69\% improvement. However, our experimental results indicate that despite these enhancements, existing LVLMs still cannot fully simulate human visual and reasoning capabilities to solve all current CAPTCHAs, which is attributed to the inherence of LVLM models. Our code and benchmark datasets are in our GitHub repository and will be available for further research. 
\end{abstract}

\keywords{Large Vision Language Models (LVLMs), CAPTCHA, Evalution, Benchmark}


\maketitle

\begin{figure}
    \centering
    \includegraphics[width=1\linewidth]{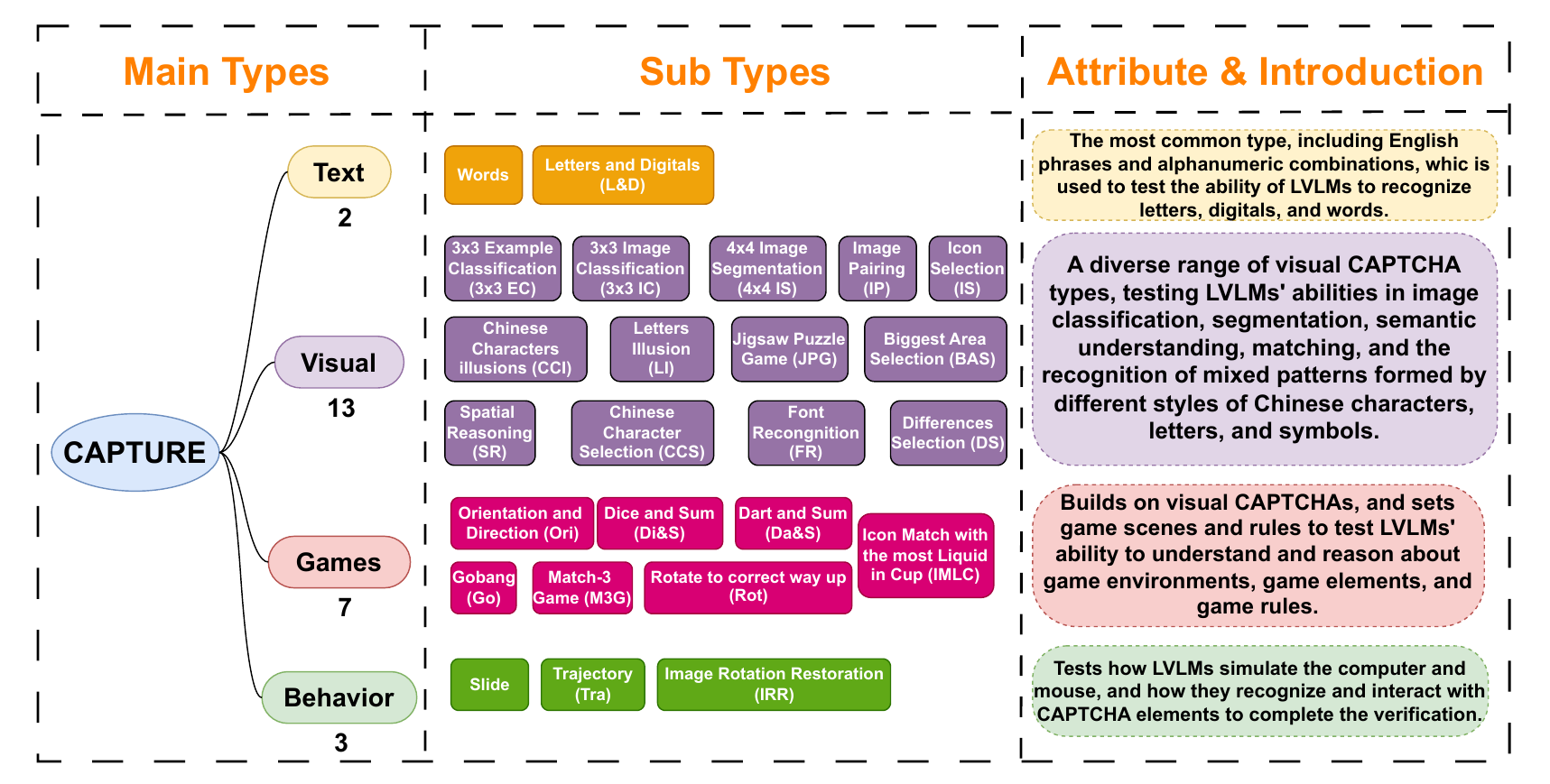}
    \caption{The \textit{CAPTURE} benchmark we designed. Correspondence between main types and sub types and attribute. The abbreviation of 25 sub-tasks are indicated in parentheses.}
    \label{fig:capture_details}
\end{figure}

\section{Introduction}



In recent years, the development of Large Language Models (LLMs)\cite{llm1-gpt4,llm2-llama,llm3-vicuna} has led to revolutionary breakthroughs in the field of Natural Language Processing (NLP)\cite{nlp-Reform-eval}. Building on these advancements, researchers are now focusing on multimodal models, particularly in computer vision. These models integrate LLMs with visual encoders\cite{blip1,blip2}and are fine-tuned using multimodal alignment methods based on generative pre-trained models\cite{Visual-instruction-tuning}. Large Visual Language Models (LVLMs)\cite{minigpt4,mplug-owl} exhibit abilities akin to human visual capabilities, such as integrating and reasoning with visual information\cite{blip2}. For instance, LVLMs can accurately analyze and interpret images\cite{minigpt4}, achieving a level of understanding comparable to that of humans' comprehensive image interpretation.

CAPTCHA (Completely Automated Public Turing Tests to Tell Computers and Humans Apart)\cite{CAPTCHA1} are widely used in verification processes on websites to differentiate between human users and malicious automated bots\cite{reCAPTCHA}. Over time, CAPTCHA systems have evolved from text-based\cite{textCAPTCHA} to image-based\cite{AsirraCAPTCHA}, and more recently, to behavior-based\cite{slider-CAPTCHAs} and reasoning-based\cite{reasoningCAPTCHA} CAPTCHA. They continue to serve as a critical security measure against malicious automated and bots. Traditional text and visual CAPTCHAs have been increasingly solved due to advancements in text detection\cite{textCAPTCHAs1,textCAPTCHAs2}, image segmentation\cite{reasoningCAPTCHA2}, and Deep Learning methods like bidirectional long short term memory (BiLSTM) network and convolutional neural network (CNN)\cite{Visual-Reasoning-CAPTCHA,reasoningCAPTCHA2}. With the advancement in LVLMs, we have a question: \textbf{\textit{Can LVLMs\%, like Silver-Bullet\footnote{In Western folklore, the silver bullet is the only weapon capable of killing werewolves, witches, or other monsters. Over time, people have used the term "silver bullet" to refer to a simple, ingenious solution for solving complex problems.}\cite{siliver-bullet}
, which emulate human visual and reasoning abilities, could further address various CAPTCHA challenges with a one-size-fits-all approach?}}

Gao et al.\cite{Visual-Reasoning-CAPTCHA,reasoningCAPTCHA2} focused primarily on text-based and image-based CAPTCHAs, attempting to solve them using Deep Learning methods. Xu et al.\cite{ViRC} compiled nearly 45,000 image CAPTCHAs from five vendors and evaluated them using the R-CNN method. Deng et al.\cite{Oedipus} analyzed five types of reasoning-based CAPTCHAs and proposed a CAPTCHA-based Domain Specific Language (DSL). They studied how to improve the success rate of LVLMs in solving reasoning CAPTCHAs through prompt words and Chain-of-Thought (CoT) techniques. Ding et al.\cite{IllusionCAPTCHA} performed an initial classification based on the design schemes of text, image, and reasoning CAPTCHAs. They conducted a systematic study and experimental analysis to evaluate the effectiveness and performance of LVLMs in solving CAPTCHAs. Their findings revealed that current CAPTCHAs are less secure than commonly perceived. Searles et al.\cite{Searles} focused on user-computer interaction and compiled ten common modern CAPTCHA forms to evaluate in user study and human-assessment level.

With the gradual improvement of web security, many reasoning-based CAPTCHA methods have emerged. Previous studies primarily focused on benchmarks and datasets related to a limited number of CAPTCHA types or methods used by mainstream websites\cite{Searles,audioCAPTCHAs1,user-friendlyCAPTCHA,JigsawPuzzle,Betterthedevilyouknow}, which have limitations in classification and comprehensiveness. Also, the studies mainly relied on traditional Deep Learning methods\cite{ViRC,Visual-Reasoning-CAPTCHA} or involved comparisons with only a few LVLMs\cite{Oedipus,IllusionCAPTCHA}, lacking a comprehensive evaluation using both open-source and closed-source LVLMs.


To address these limitations and keep pace with the dynamic development of CAPTCHAs, there is an urgent need for a new, comprehensive CAPTCHA evaluation benchmark tailored for LVLMs. Based on the analysis of the aforementioned methods, we identify the following characteristics that a comprehensive CAPTCHA evaluation benchmark for LVLMs should possess:

\begin{enumerate}
    \item Comprehensiveness: The benchmark should encompass a wide range of commonly encountered CAPTCHAs, as well as the latest proposed secure CAPTCHA paradigms.
    \item Novelty of Data Sources: The benchmark should avoid direct reliance on existing public datasets to minimize the risk of data leakage and maintain its novelty and timeliness.
    \item Consistency with Real-World Scenarios: The CAPTCHA dataset should be sourced directly from real-world online environments and web applications rather from synthesized or artificial data, reflecting problems that are challenging for machines but easy for humans. This ensures an accurate evaluation of LVLMs' ability to solve diverse and complex CAPTCHA problems.
    \item Balanced Difficulty Levels: CAPTCHA problems vary in difficulty for human users. Therefore, the dataset should categorize CAPTCHA types by difficulty, enabling a more effective assessment of whether LVLMs can recognize, reason, and solve CAPTCHAs at varying levels of complexity.
\end{enumerate}


Guided by the characteristics, we propose a comprehensive benchmark for evaluating LVLMs in solving CAPTCHAs in complex real-world scenarios, named \textit{CAPTURE} (\textbf{CAP}TCHA for \textbf{T}esting \textbf{U}nder \textbf{R}eal-world \textbf{E}xperiments). Figure \ref{fig:capture_details} shows the details of \textit{CAPTURE}. Additionally, we have designed a comprehensive evaluation benchmark dataset. The main contributions of this paper are as follows:

\begin{itemize}
    \item We reviewed previous CAPTCHA evaluation benchmark schemes and proposed four key characteristics essential for a robust benchmark.
    \item We propose the \textit{CAPTURE} Benchmark, which is categorized by CAPTCHA types, includes 4 main types and 25 subtypes. It enabling a comprehensive evaluation of LVLMs.s
    \item We evaluate commonly LVLMs including both open-source and closed-source using \textit{CAPTURE} and we propose a two-stage framework \textbf{CRRD} aiming at further enhancing the ability in solving complex CAPTCHAs.
\end{itemize}

\section{Related Work}

\subsection{CAPTCHA Resolving}
Existing CAPTCHAs can be classified into several types: text-based, image-based, audio-based, video-based and behavioral CAPTCHAs\cite{Searles}. In 2014, text recognition algorithms were able to solve text-based CAPTCHAs with over 99\% accuracy, significantly surpassing the human accuracy rate of 33\%.\cite{googlenet} Researchers have gradually developed effective methods to solve behavioral CAPTCHAs. Andreas et al.\cite{Breakingrecaptchav2} used the YOLO model to identify and break Google’s reCAPTCHA v2\cite{reCAPTCHAv2} with nearly 100\% accuracy. Xu et al.\cite{ViRC} used Graph Neural Networks (GNNs) to solve several types of image-based behavioral CAPTCHAs. Benjamin\cite{WindMouse} proposed the Windmouse method, which employs computer simulations to model human control of mouse movement trajectories. Gao et al.\cite{reasoningCAPTCHA2} employed the Faster R-CNN method to break five commonly used sliding CAPTCHAs with a success rate exceeding 87.5\%, and combined BiLSTM with Faster R-CNN to tackle various complex text and image-based CAPTCHAs.

With the rapid development and widespread usage of LVLMs, researchers have started applying them to solve CAPTCHA problems. Deng et al.\cite{Oedipus} proposed a two-stage framework consisting of offline generation and online solving, introducing the CAPTCHA Domain Specific Language (DSL) paradigm. They used GPT-4V\cite{gpt4v} and Gemini\cite{gemini} to solve certain reasoning-based CAPTCHAs. Ding et al.\cite{IllusionCAPTCHA} tested GPT-4o\cite{gpt-4o} and Gemini 1.5 Pro\cite{gemini1.5} on text, image, and reasoning-based CAPTCHAs. Using prompt engineering and CoT\cite{cot} techniques, they improved the solving accuracy and proposed a type of novel CAPTCHA - Illusion CAPTCHA.

However, the solving methods mentioned above that using DNNs are implemented in highly controlled environments, which may not reflect their vision performance in real-world settings. Using LVLMs to attempt solving CAPTCHA solutions, the benchmarks and datasets are not sufficiently comprehensive, and the LVLMs lack diversity, failing to accurately reflect their true performance and capabilities in solving CAPTCHAs.

\subsection{Large Vision Language Models (LVLMs)}

Since LLMs are leading the development in the NLP field, researchers have further combined visual encoders with LLMs, aligning multimodal models through generative pretraining\cite{nlp-Reform-eval}. This enables LVLMs to understand and reason with visual information, better simulating human thinking, reasoning, and analysis in visual data like CAPTCHAs. Compared to traditional deep learning models, LVLMs exhibit notable abilities, such as OCR\cite{ocrbench}, meme words and sentences understanding\cite{minigpt4}, and commonsense reasoning with visual images\cite{blip2}. The deep thinking functionality, first proposed by DeepSeek\cite{deepseekr1}, has been widely applied to LVLMs, further enhancing their ability to reason with visual information. 

On the other hand, based on our recent research, the lack of suitable comprehensive evaluation benchmarks and datasets in CAPTCHA reasoning and solving hindered in-depth quantitative analysis and comparison of LVLMs in this area. To the best of our knowledge, there are few comprehensive CAPTCHA benchmarks or datasets suitable for evaluating LVLMs' capabilities, limiting quantitative analysis and comparison of these models.

\begin{figure}[h]
    \centering
    \includegraphics[width=1\linewidth]{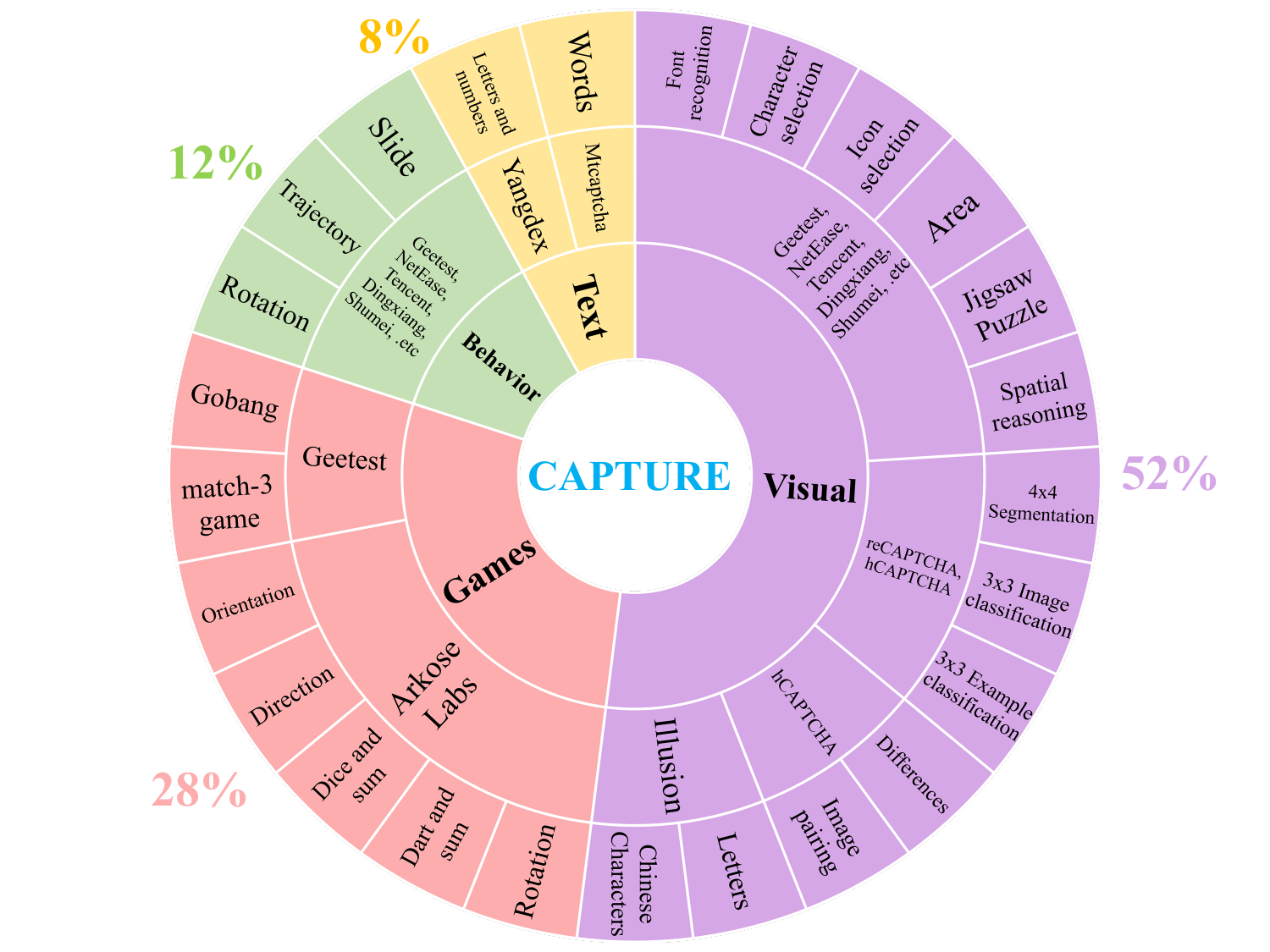}
    \caption{The classification and CAPTCHA types in \textit{CAPTURE} benchmark}
    \label{fig:CAPTURE types}
\end{figure}

\begin{figure*}
    \centering
    \includegraphics[width=1\linewidth]{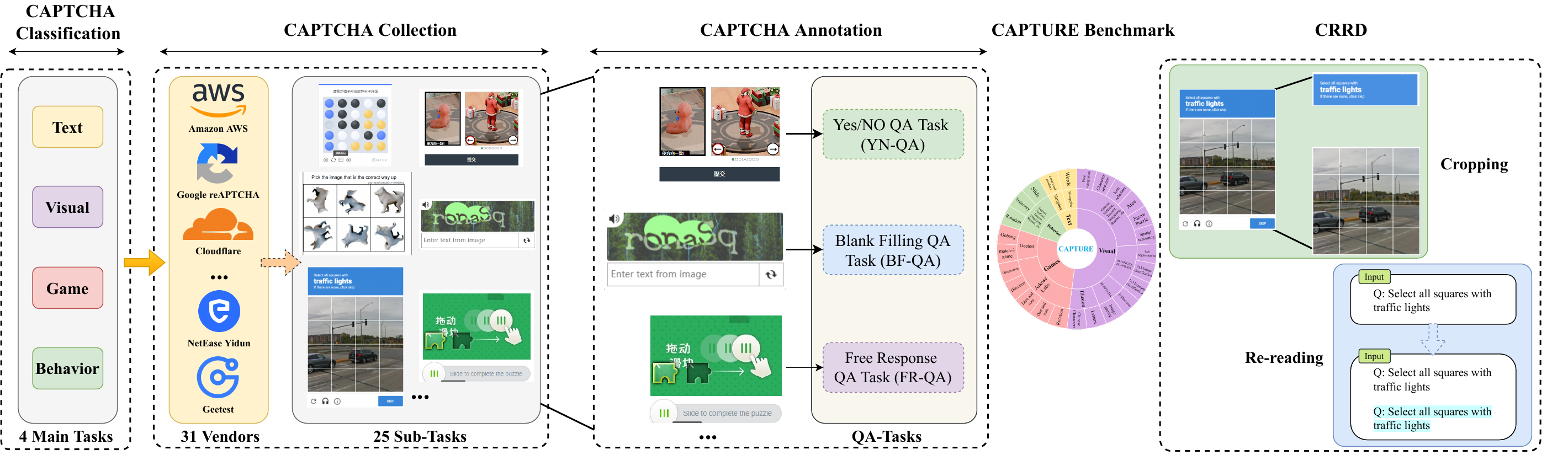}
    \caption{Building Process of CAPTURE Benchmark.}
    \label{fig:pipeline}
\end{figure*}

\begin{table*}[t]
 \caption{The details of previous CAPTCHA-related studies and datasets (in the past 5 years) and our work.}
  \label{tab:study}
\begin{tabular}{lcccclcc}
\toprule
\multicolumn{1}{c}{\multirow{2}{*}{\textbf{Dataset}}} &
  \multicolumn{4}{c}{CAPTCHA Types} &
  \multicolumn{1}{c}{\multirow{2}{*}{\textbf{CAPTCHA Source}}} &
  \multicolumn{1}{c}{\multirow{2}{*}{\textbf{CAPTCHAs}}} &
  \multicolumn{1}{c}{\multirow{2}{*}{\textbf{Year}}} \\ \cline{2-5}
\multicolumn{1}{c}{} &
  Text &
  Images &
  Games &
  Behavior &
  \multicolumn{1}{c}{} &
  \multicolumn{1}{c}{} &
  \multicolumn{1}{c}{} \\ 
  \midrule
Gao et.al \cite{Visual-Reasoning-CAPTCHA} &
   &
  \checkmark &
   &
   &
  Major websites \cite{Tencent}, \cite{geetest}, \cite{netease}, \cite{shumei} &
  13,500 &
  2021 \\
Searles et.al \cite{Searles} &
  \checkmark &
  \checkmark &
  \checkmark &
  \checkmark &
  Alexa\cite{Alexa} &
  14,000 &
  2023 \\
Xu et.al \cite{ViRC} &
   &
  \checkmark &
   &
   &
  Major websites \cite{Tencent}, \cite{geetest}, \cite{netease}, \cite{shumei} &
  40,300 &
  2023 \\
Deng et.al \cite{Oedipus} &
   &
  \checkmark &
  \checkmark &
   &
  Major websites \cite{ArkoseLabs} \cite{geetest} \cite{netease} &
  50 &
  2024 \\
Ding et.al \cite{IllusionCAPTCHA} &
  \checkmark &
  \checkmark &
  \checkmark &
   &
  Major websites \cite{reCAPTCHAv2} \cite{ArkoseLabs} \cite{hCaptcha} \cite{geetest} \cite{netease} &
  330 &
  2025 \\
\textbf{Ours} &
  \checkmark &
  \checkmark &
  \checkmark &
  \checkmark &
  \textbf{31 CAPTCHA vendors} &
  \textbf{61,464} &
  \textbf{2025} \\ 
  \bottomrule
\end{tabular}%
\end{table*}



\begin{table}[t!]
\caption{CAPTCHA usage distribution on the Internet}
\label{tab:CAPTCHA distribution}
\resizebox{\columnwidth}{!}{
\begin{tabular}{cccccc}
\toprule
\multicolumn{3}{c}{\textbf{World}}       & \multicolumn{3}{c}{\textbf{China}}     \\ 
Technology & Websites   & \%    & Technology & Websites & \%    \\ \midrule
reCAPTCHA\cite{reCAPTCHAv2}  & 10,171,688 & 89.74 & reCAPTCHA  & 22,173   & 77.34 \\
hCaptcha\cite{hCaptcha}   & 811,395    & 7.16  & hCaptcha   & 1,791    & 6.25  \\
Cloudflare\cite{CloudFlare} & 222,815    & 1.97  & Geetest\cite{geetest}    & 1,447    & 5.05  \\
Others     & 128,414    & 1.13  & Others     & 6,980    & 11.36 \\ 
Total      & 11,334,312 & 100   & Total      & 32,391   & 100 \\ \bottomrule
\end{tabular}%
}
\end{table}

\subsection{CAPTCHA Datasets}

Existing open-source CAPTCHA datasets are primarily sourced from repositories in such as GitHub\footnote{https://github.com/search?q=captcha\&type=repositories} and Huggingface\footnote{https://huggingface.co/datasets?search=captcha}, focus on traditional text-based and image-based CAPTCHAs. Andreas et al.\cite{Breakingrecaptchav2} focused solely on image classification and segmentation tasks from Google reCAPTCHA v2\cite{reCAPTCHAv2} for their research. Gao et al. initially focused on character-based\cite{Text-based-CAPTCHA}, sliding\cite{slider-CAPTCHAs}, and image-based\cite{Visual-Reasoning-CAPTCHA} CAPTCHAs, emphasizing the collection and organization of image-based CAPTCHAs from several Chinese CAPTCHA vendors (Geetest\cite{geetest}, NetEase\cite{netease}, Dingxiang\cite{dingxiang}, and Shumei\cite{shumei}). They subsequently conducted a classification study of text-based and image-based CAPTCHAs across three dimensions. Xu et al.\cite{ViRC} constructed the ViRC dataset, containing over 40,000 manually labeled image-based CAPTCHA images for deep learning. Deng et al.\cite{Oedipus} and Ding et al.\cite{IllusionCAPTCHA}, further collected CAPTCHA schemes from internationally representative CAPTCHA vendors. Searles et al.\cite{Searles} compiled CAPTCHAs from globally well-known websites ranked by Alexa\cite{Alexa}. Then they conducted experiments and evaluations from the dimensions of the time required for humans to solve CAPTCHAs, their preferences for different types of CAPTCHAs, and the context settings of CAPTCHAs. 

As shown in Table \ref{tab:CAPTCHA distribution}, Google reCAPTCHA are widely deployed in most of the websites. This denotes that for the websites selected based on the ranking, most of them are likely to use Google reCAPTCHA for  verification. Also, previous studies focused on the CAPTCHA types they primarily aimed to solved, mostly approaching the problem from the perspective of solving one or a few types of CAPTCHA. They then proposed algorithms and benchmarks, which accompanied with datasets and adapted to these algorithms. Thus, the CAPTCHA datasets and benchmarks mentioned above are not sufficiently comprehensive, Specifically, in the realm of LVLMs' evaluation, there are few comprehensive benchmarks or datasets to evlauate the ability in solving CAPTCHA problems.

\section{Benchmark Designing}
In this Section, we will describe in details how we construct Benchmark. Notably, our Benchmark primarily serves as a test set, with no dedicated training or validation sets, focusing on evaluating the reasoning and problem-solving capabilities of LVLMs in real-world scenarios. The buliding process of our benchmark is shown in Figure \ref{fig:pipeline}. The brief introductions of previous CAPTCHA-related studies and our works are shown in Table \ref{tab:study}. And the full Table is shown in Table \ref{tab:vendors} in Appendix.

\subsection{CAPTCHA Classification}
Generally, based on the form of presentation and resolving methods, CAPTCHAs are divided into several main types: text, image, games, behavior, audio, video, .etc. In this paper, we mainly focus on whether LVLMs could simulate human visual and reasoning abilities to solve various of CAPTCHAs. Therefore, the types discussed here only include the \textbf{text}, \textbf{image}, \textbf{games} and \textbf{behavior}. Inspired by Searles et.al\cite{Searles} we further divided these 4 main tasks into 25 sub-tasks. The classification is shown in the Figure \ref{fig:CAPTURE types}\footnote{The details of each main tasks and sub-tasks are shown in Supplementary Material.}.  We collect and organize the CAPTCHAs according to the classification, and then form our dataset.

\subsection{CAPTCHA Collection}

Searles et al.\cite{Searles} selected the top 200 websites from Amazon Alexa\cite{Alexa} "Top Website List" that  based on the popularity and collected the CAPTCHAs. However, most of websites deployed the reCAPTCHA from Google. Table \ref{tab:CAPTCHA distribution} shows the latest CAPTCHA usage distribution in the world and in China.

Thus, We investigate CAPTCHAs based on their vendors that specialized in security services. Concerning the languages that would change the caption and content of CAPTCHAs (e.g. text-based CAPTCHAs), we mainly focus on English and Chinese vendors\footnote{Table \ref{tab:vendors} in Supplementary Material shows the vendors we investigated.}.  We directly accessed the CAPTCHA vendor’s homepage and the user registration pages of websites using the CAPTCHAs. Using Selenium\footnote{https://www.selenium.dev/} and RPA\cite{RPA} (Robotic Process Automation) technology, we retrieve and downloade the CAPTCHA from these pages. By using the ready-made CAPTCHAs from these sources, we guarantee the authenticity and validity of the CAPTCHA.

\subsection{CAPTCHA Annotation}

Different types of CAPTCHA have different representations, and the methods for solving them vary. As a result, manually annotating each CAPTCHA image is costly. Inspired by Shen et.al \cite{shen2023chatgpt}, we design 3 QA Task methods: Yes/NO QA Task (YN-QA), Blank Filling QA Task (BF-QA), and Free Response QA Task (FR-QA) for CAPTCHA Label annotation. As shown in Table \ref{tab:qa_formats}, these three annotation methods can be applied to the 25 types of CAPTCHA in \textit{CAPTURE}.

\textbf{Label annotation:} We use LLaVA-1.6\cite{llava} as the LVLM for annotation. Different annotation tasks are used depending on the CAPTCHA type. In Table \ref{tab:qa_formats}, we take 3 CAPTHCA demo as examples. ArkoseLabs FunCaptcha is a type of Game CAPTCHA, and we use YN-QA Task to directly judge the CAPTCHA, which can be seen as a binary classification task. MTcaptcha is a text-based CAPTCHA, which directly outputs the text content and uses BF-QA. The differnet task of Dingxiang is considered as FR-QA, where the LVLMs needs to identify a distinct feature in the image and provide a detailed description of it.

\textbf{Label verification and extention:} During the annotation process, we noticed that LLAVA performed poorly on some CAPTCHA types. Solving CAPTCHAs is relatively simple for humans, so we invited 25 testers to manually verify the annotated CAPTCHAs, revise any errors in the annotations, and supplement incomplete descriptions. Finally, we supplement with information such as the CAPTCHA classification and the source vendor, forming complete label information for each CAPTCHA.

\begin{table}[h]
\footnotesize
\caption{The examples of 3 common QA tasks.}
\label{tab:qa_formats}
\begin{tabular}{p{.15\linewidth}|p{.80\linewidth}}
\toprule
\multicolumn{2}{c}{\cellcolor[HTML]{D9EAD3}{\textbf{Yes/NO QA Task (YN-QA)}}}   \\
\midrule
Context     &  Sum the dice and match the number on the left. \\
Source & ArkoseLabs FunCaptcha \\
& \includegraphics[width=0.4\linewidth]{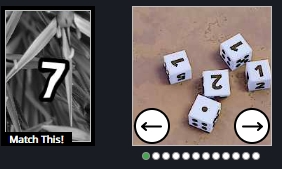} \\ 
Question & Are the numbers on the left side equal to the sum of the dice points on the right side? \yn{\textbf{Yes or N}o?}  \\
Answer & \textbf{\yn{NO}} \\ 
\toprule
\multicolumn{2}{c}{\cellcolor[HTML]{CFE2F3}\textbf{Blank Filling QA Task (BF-QA)}}   \\
\midrule
Context  & Enter texts from the image (Case Sensitive) \\
Source & MTcaptcha \\
& \includegraphics[width=0.4\linewidth]{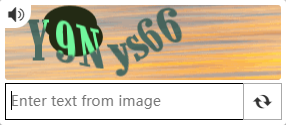} \\ 
Question   & The texts in the image is: \extractive{(   )}   \\
Answer  & The texts in the image is: (\textbf{\extractive{Y9Nys66}})  \\
\toprule
\multicolumn{2}{c}{\cellcolor[HTML]{E1D5E7}\textbf{Free Response QA Task (FR-QA)}} \\
\midrule
Context  & Choose the pattern that is different from the others. \\
Source & Dingxiang \\
& \includegraphics[width=0.4\linewidth]{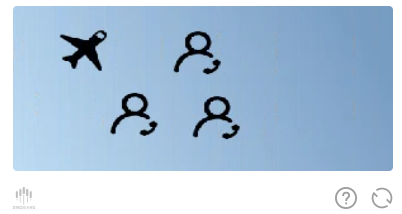} \\
Question & Which pattern should you choose? What is the pattern? Where is the location? \\
Answer & There are three identical character patterns and one airplane pattern. The \abs{airplane} is different from the others. The position is \abs{the first one from the left.} \\
\bottomrule
\end{tabular}

\end{table}

\subsection{Data Summarization}

Based on feedback from the testers during the CAPTCHA labeling process, we observe that some CAPTCHAs are more challenging to solve, often requiring multiple attempts to achieve the correct result. To further investigate, the testers solve 25 types of CAPTCHAs under real-world conditions and the results compiled the corresponding human success rates and we will analyse in Section \ref{section:Experiment}. We then integrate CAPTCHAs with similar verification methods from different vendors. The data includes efficiency on defensing against automated bots and user experience, which allowing for a comparison of the effectiveness of different verification methods. The results show in Table \ref{tab:basic} (Human Acc) and reveal varying success rates across different type of CAPTCHAs, help define the difficulty. This ensures that the proposed benchmark balances CAPTCHAs of varying difficulty, providing a more accurate assessment of LVLMs' capabilities.

\subsection{Data Statistics}
In summary, our dataset includes 4 main tasks, and then into 25 sub-tasks, with a total of 61,464 CAPTCHAs. The data and proportions for each main task are shown in Table \ref{tab:Data statistics}.

\begin{table}[htbp]

\centering
\caption{Data statistics}
\label{tab:Data statistics}
\begin{tabular}{cccc}
\toprule
\textbf{Main Type} & \textbf{Sub Type} & \textbf{Count} & \textbf{\%} \\ \midrule
Text               & 2                 & 4,917          & 0.08        \\
Visual             & 13                & 31,961         & 0.52        \\
Games              & 7                 & 17,209         & 0.28        \\
Behavior           & 3                 & 7,377          & 0.12        \\
\textbf{Total}              & \textbf{25}                & \textbf{61464}          & 1.00           \\ \bottomrule
\end{tabular}%
\end{table}

\section{Experiments} \label{section:Experiment}
In this Section, first we propose several Research Questions for the experiments. Then we provide descriptions of our experimental implementations and evaluation metrics. We initially analyze and explain the experimental results and  propose a two-stage framework, \textbf{CRRD} (\textbf{C}ropping, \textbf{R}e-\textbf{R}eading and \textbf{D}escribing), aimed at improving LVLMs' performance in solving CAPTCHAs from our dataset.

\begin{table}[htbp]
\caption{The full names with abbreviations for sub-tasks and the CAPTCHA numbers in each task. Different colors indicate the different main tasks, corresponding to Figure \ref{fig:capture_details} and Figure \ref{fig:CAPTURE types}}
\label{full-abbr}

\resizebox{\columnwidth}{!}{%
\begin{tabular}{ccc|ccc}
\hline
\multicolumn{1}{c}{\textbf{Sub-Tasks}} & \textbf{Abbr.} & \textbf{\# CAPTCHA} & \multicolumn{1}{c}{\textbf{Sub-Tasks}} & \textbf{Abbr.} & \textbf{\# CAPTCHA} \\ \hline
\rowcolor[HTML]{E1D5E7} 
3x3 Example classification &
  3x3 EC &
  4417 &
  \cellcolor[HTML]{FFF2CC}Words &
  \cellcolor[HTML]{FFF2CC}Words &
  \cellcolor[HTML]{FFF2CC}2462 \\
\rowcolor[HTML]{E1D5E7} 
3x3 Image classification &
  3x3 IC &
  4021 &
  \cellcolor[HTML]{FFF2CC}Letters and Digitals &
  \cellcolor[HTML]{FFF2CC}L\&D &
  \cellcolor[HTML]{FFF2CC}2455 \\
\rowcolor[HTML]{E1D5E7} 
4x4 Image Segmentation &
  4x4 IS &
  4452 &
  \cellcolor[HTML]{FFCCC9}Slide &
  \cellcolor[HTML]{FFCCC9}Slide &
  \cellcolor[HTML]{FFCCC9}10209 \\
\rowcolor[HTML]{E1D5E7} 
Chinese Characters Illusion &
  CCI &
  92 &
  \cellcolor[HTML]{FFCCC9}Trajectory &
  \cellcolor[HTML]{FFCCC9}Tra &
  \cellcolor[HTML]{FFCCC9}3247 \\
\rowcolor[HTML]{E1D5E7} 
Letters  Illusion &
  LI &
  26 &
  \cellcolor[HTML]{FFCCC9}Image Rotation Restoration &
  \cellcolor[HTML]{FFCCC9}IRR &
  \cellcolor[HTML]{FFCCC9}3753 \\
\rowcolor[HTML]{E1D5E7} 
Image pairing &
  IP &
  2891 &
  \cellcolor[HTML]{E2EFDA}Orientation \& Direction &
  \cellcolor[HTML]{E2EFDA}O\&D &
  \cellcolor[HTML]{E2EFDA}1052 \\
\rowcolor[HTML]{E1D5E7} 
Differences Selection &
  DS &
  1289 &
  \cellcolor[HTML]{E2EFDA}Icon Match &
  \cellcolor[HTML]{E2EFDA}IM &
  \cellcolor[HTML]{E2EFDA}1543 \\
\rowcolor[HTML]{E1D5E7} 
Spatial reasoning &
  SR &
  2302 &
  \cellcolor[HTML]{E2EFDA}Dice and sum &
  \cellcolor[HTML]{E2EFDA}Di\&M &
  \cellcolor[HTML]{E2EFDA}384 \\
\rowcolor[HTML]{E1D5E7} 
Jigsaw Puzzle &
  JP &
  2871 &
  \cellcolor[HTML]{E2EFDA}Dart and sum &
  \cellcolor[HTML]{E2EFDA}Da\&M &
  \cellcolor[HTML]{E2EFDA}862 \\
\rowcolor[HTML]{E1D5E7} 
Biggest Area Selection &
  BAS &
  1757 &
  \cellcolor[HTML]{E2EFDA}Rotate to correct way up &
  \cellcolor[HTML]{E2EFDA}Rot &
  \cellcolor[HTML]{E2EFDA}1159 \\
\rowcolor[HTML]{E1D5E7} 
Icon selection &
  IS &
  3284 &
  \cellcolor[HTML]{E2EFDA}Gobang &
  \cellcolor[HTML]{E2EFDA}Go &
  \cellcolor[HTML]{E2EFDA}1631 \\
\rowcolor[HTML]{E1D5E7} 
Chinese Character selection &
  CS &
  1824 &
  \cellcolor[HTML]{E2EFDA}Match-3 Game &
  \cellcolor[HTML]{E2EFDA}M3G &
  \cellcolor[HTML]{E2EFDA}746 \\
\cellcolor[HTML]{E1D5E7}Font recognition &
  \cellcolor[HTML]{E1D5E7}FR &
  \cellcolor[HTML]{E1D5E7}2735 &
   &
  \multicolumn{1}{l}{} &
  \multicolumn{1}{l}{} \\ \hline
\end{tabular}%
}
\end{table}

\subsection{Research Questions}
In our experiments, we first propose 2 Research Questions for further analysis.

\textbf{RQ1: Could LVLMs solve all types of CAPTCHAs?} To address this, we classify LVLMs based on whether they are open-source or closed-source and perform a comprehensive evaluation of both categories.

\textbf{RQ2: How about input methods and prompt engineering techniques enhance LVLMs' ability to solve CAPTCHAs?} To further explore the capabilities of LVLMs, we propose a two-stage framework to process input images and design prompts, aiming to better enhance LVLMs' reasoning ability and analyze whether LVLMs can solve CAPTCHA problems.

\subsection{Implementations}
All the CAPTCHA we tested are sourced from the \textit{CAPTURE} dataset. We tested popular LVLMs on the market based on LLM Leaderboard in Huggingface\footnote{https://huggingface.co/spaces/open-llm-leaderboard/open\_llm\_leaderboard} and API vendor\footnote{https://api.v3.cm/}. We classified LVLMs into open-source and closed-source categories. For closed-source LVLMs, we request their responses via API calls. For open-source LVLMs, we deployed them on NVIDIA 4090 GPUs for the experiments.

\subsection{Evaluation Metrics}

In designing the evaluation metrics, we adopt different evaluation methods based on the dataset labeling approach for three different types of verification: Yes/NO QA Task (YN-QA), Blank Filling QA Task (BF-QA), and Free Response QA Task (FR-QA).

For CAPTCHAs labeled with the YN-QA and BF-QA methods, we directly compare the output result with the label to check for consistency. For CAPTCHAs labeled with the FR-QA method, The evaluation metric is the word-level accuracy rate: the proportion of the output that fully presents the true values of words.


Among these three types of response methods, the task of the model is to generate a reasonable answer based on the question $q$ and the vision context $v$. Although these three QA-tasks are different from each other, they can all be unified. We denote $P_{\theta}(r | v,q)$ as the probability of model $M$ generates response $r$, that $r$ represents the final response from $M$. Whether it is the answer of YN-QA, BF-QA or FR-QA. Specifically, as for YN-QA task, $M$ generates a binary $r$ Yes/No. As for BF-QA task, $M$ generates a concrete string, words or phrase $r$. As for FR-QA, $M$ generates Generate a longer text as $r$. Formally, the answer prediction can be determined by the generation likelihood predicted by the evaluated model:

\begin{equation}
\hat{r} = \arg\max_{r} P_\theta(r | v, q) 
\end{equation}

Extended generation probability refers to the step-by-step generation process of generating a sequence of response. For all three types of QA-tasks, the response is a sequence (usually a token of length 1 in YN-QA and a longer sequence in BF-QA and FR-QA). We can generate tokens step-by-step in an autoregressive manner and calculate the generation probability of each token:

\begin{equation}
\hat{r} = \arg\max_{r} P_\theta(r | v, q) = \arg\max_{r} \sum_{t=1}^{T_r} P_\theta(r_t | v, q, r_{<t}) 
\end{equation}

While $r_t$ is the t-th token in the generated response sequence. $r_{<t}$ is the sequence of all previously generated tokens (i.e., the context of the model). $T$ is the length of the answer sequence. The specific forms of $r$ and T for different types of tasks are as follows: 

\begin{itemize}
    \item \textbf{YN-QA:} $r$ has only two possible options "Yes" or "No", with $T=1$, where the model generates a single word.
    \item \textbf{BF-QA:} $r$ is typically a single word or a phrase, with $T \geq 1$.
    \item \textbf{FR-QA:} $r$ is a longer text, with $T > 1$, where each token is part of the generated answer until the end-of-sequence token is produced.
\end{itemize}

\begin{table*}[h]
\footnotesize
\centering
\caption{LVLMs evaluation using basic prompt.}
\label{tab:basic}
\begin{threeparttable}
\begin{tabular}{cccccccccccccc}
\toprule
 &
   &
  \multicolumn{6}{c}{\textbf{Closed Sourced LVLMs}} &
  \multicolumn{5}{c}{\textbf{Open Sourced LVLMs}} &
   \\ \cline{3-13}
\multirow{-2}{*}{\textbf{\begin{tabular}[c]{@{}c@{}}Sub Types \\ (Abbr.)\end{tabular}}} &
  \multirow{-2}{*}{\textbf{\begin{tabular}[c]{@{}c@{}}Human\\ Acc\end{tabular}}} &
  \begin{tabular}[c]{@{}c@{}}GPT\\ 4o\end{tabular} &
  \begin{tabular}[c]{@{}c@{}}GPT-4o\\ mini\end{tabular} &
  \begin{tabular}[c]{@{}c@{}}Claude\\ 3.5\end{tabular} &
  \begin{tabular}[c]{@{}c@{}}Doubao\\ 1.5\end{tabular} &
  \begin{tabular}[c]{@{}c@{}}Gemini\\ 1.5\end{tabular} &
  \begin{tabular}[c]{@{}c@{}}Gemini\\ 2.0\end{tabular} &
  \begin{tabular}[c]{@{}c@{}}GLM\\ 4v\end{tabular} &
  Grok2 &
  \begin{tabular}[c]{@{}c@{}}Qwen\\ 2.5\end{tabular} &
  QVQ &
  \begin{tabular}[c]{@{}c@{}}Deep\\ Seek\end{tabular} &
  \multirow{-2}{*}{\textbf{\begin{tabular}[c]{@{}c@{}}LVLMs\\ Avg\end{tabular}}} \\ \midrule
\rowcolor[HTML]{FFF2CC} 
L\&D &
  \cellcolor[HTML]{EFEFEF}0.8738 &
  0.6190 &
  0.6190 &
  0.4762 &
  0.5714 &
  0.5714 &
  \textbf{0.6667} &
  0.6190 &
  0.5714 &
  0.5714 &
  0.3810 &
  0.5714 &
  0.5671 \\
\rowcolor[HTML]{FFF2CC} 
Words &
  \cellcolor[HTML]{EFEFEF}0.9980 &
  \textbf{0.9534} &
  {\ul 0.8571} &
  0.7143 &
  0.8571 &
  {\ul 0.9504} &
  0.9524 &
  0.6667 &
  {\ul 0.8571} &
  0.8571 &
  0.6667 &
  {\ul 0.9048} &
  {\ul 0.8398} \\
\rowcolor[HTML]{E1D5E7} 
3x3 EC &
  \cellcolor[HTML]{EFEFEF}0.9415 &
  0.0526 &
  0.0000 &
  0.0000 &
  0.0526 &
  0.1842 &
  \textbf{0.2632} &
  0.0263 &
  0.1316 &
  0.0789 &
  0.0263 &
  0.0263 &
  \textbf{0.0766} \\
\rowcolor[HTML]{E1D5E7} 
3x3 IC &
  \cellcolor[HTML]{EFEFEF}0.9363 &
  0.5000 &
  0.1053 &
  0.5000 &
  0.5526 &
  0.6842 &
  {\ul 0.9749} &
  0.1316 &
  0.7105 &
  0.6842 &
  0.4474 &
  0.0789 &
  0.4928 \\
\rowcolor[HTML]{E1D5E7} 
4x4 IS &
  \cellcolor[HTML]{EFEFEF}0.9650 &
  0.0263 &
  0.2368 &
  0.2632 &
  \textbf{0.3158} &
  0.2895 &
  0.2368 &
  0.2895 &
  0.2895 &
  \textbf{0.3158} &
  0.2105 &
  0.2105 &
  0.2440 \\
\rowcolor[HTML]{E1D5E7} 
CCI &
  \cellcolor[HTML]{EFEFEF}0.9952 &
  0.6316 &
  0.5263 &
  0.5526 &
  0.4962 &
  \textbf{0.7368} &
  0.6842 &
  0.4211 &
  0.5789 &
  0.1579 &
  0.1579 &
  0.4211 &
  0.4877 \\
\rowcolor[HTML]{E1D5E7} 
LI &
  \cellcolor[HTML]{EFEFEF}0.9993 &
  0.0188 &
  0.0020 &
  0.0050 &
  0.0040 &
  0.0030 &
  0.0020 &
  0.0119 &
  0.0139 &
  0.0168 &
  0.0020 &
  0.0178 &
  0.0088 \\
\rowcolor[HTML]{E1D5E7} 
IP &
  \cellcolor[HTML]{EFEFEF}0.9761 &
  0.4474 &
  0.0263 &
  0.1579 &
  \textbf{0.8421} &
  \textbf{0.8421} &
  \textbf{0.8421} &
  0.0526 &
  0.3421 &
  0.2895 &
  0.1316 &
  0.1842 &
  0.3780 \\
\rowcolor[HTML]{E1D5E7} 
DS &
  \cellcolor[HTML]{EFEFEF}0.9601 &
  0.7368 &
  0.3684 &
  0.5789 &
  \textbf{0.7895} &
  0.7368 &
  0.7368 &
  {\ul 0.6842} &
  0.2632 &
  0.3158 &
  0.3684 &
  0.5263 &
  0.5550 \\
\rowcolor[HTML]{E1D5E7} 
SR &
  \cellcolor[HTML]{EFEFEF}0.9130 &
  {\ul 0.9505} &
  0.6667 &
  0.6190 &
  0.8571 &
  0.7619 &
  0.9524 &
  0.5238 &
  0.1905 &
  0.9048 &
  0.8571 &
  {\ul 0.9048} &
  0.7489 \\
\rowcolor[HTML]{E1D5E7} 
JP &
  \cellcolor[HTML]{EFEFEF}0.8782 &
  0.0000 &
  0.0000 &
  0.0000 &
  0.0000 &
  \textbf{0.0526} &
  \textbf{0.0526} &
  0.0000 &
  0.0000 &
  0.0000 &
  0.0000 &
  0.0000 &
  0.0096 \\
\rowcolor[HTML]{E1D5E7} 
BAS &
  \cellcolor[HTML]{EFEFEF}0.9849 &
  0.1053 &
  0.0263 &
  0.0263 &
  \textbf{0.1579} &
  0.0789 &
  0.1316 &
  0.0263 &
  0.0263 &
  0.0263 &
  0.0000 &
  0.0263 &
  0.0574 \\
\rowcolor[HTML]{E1D5E7} 
IS &
  \cellcolor[HTML]{EFEFEF}0.9582 &
  0.3158 &
  0.0263 &
  0.3158 &
  0.4737 &
  0.1842 &
  \textbf{0.5000} &
  0.0263 &
  0.1842 &
  0.2895 &
  0.0263 &
  0.0263 &
  0.2153 \\
\rowcolor[HTML]{E1D5E7} 
CCS &
  \cellcolor[HTML]{EFEFEF}0.9613 &
  0.0000 &
  0.0000 &
  0.0263 &
  0.0000 &
  0.0000 &
  \textbf{0.3158} &
  0.0000 &
  0.0000 &
  0.0000 &
  0.0263 &
  0.0263 &
  0.0359 \\
\rowcolor[HTML]{E1D5E7} 
FR &
  \cellcolor[HTML]{EFEFEF}0.9890 &
  0.4474 &
  0.0526 &
  0.3158 &
  0.3421 &
  0.7105 &
  \textbf{0.7895} &
  0.0000 &
  0.0789 &
  0.3158 &
  0.1579 &
  0.2895 &
  0.3182 \\
\rowcolor[HTML]{FFCCC9} 
\cellcolor[HTML]{F8CECC}Slide &
  \cellcolor[HTML]{EFEFEF}0.9973 &
  0.1053 &
  0.0000 &
  0.2632 &
  0.0526 &
  0.2368 &
  \textbf{0.2895} &
  0.0526 &
  0.1053 &
  0.0526 &
  0.0000 &
  0.0263 &
  0.1077 \\
\rowcolor[HTML]{FFCCC9} 
\cellcolor[HTML]{F8CECC}Tra &
  \cellcolor[HTML]{EFEFEF}0.9733 &
  0.5000 &
  0.4737 &
  0.5789 &
  0.4737 &
  0.5263 &
  \textbf{0.6579} &
  0.3947 &
  0.1842 &
  0.4211 &
  0.2632 &
  0.3947 &
  0.4426 \\
\rowcolor[HTML]{FFCCC9} 
\cellcolor[HTML]{F8CECC}IRR &
  \cellcolor[HTML]{EFEFEF}0.9570 &
  0.0000 &
  0.0000 &
  0.2895 &
  0.0263 &
  0.0789 &
  \textbf{0.3158} &
  0.0000 &
  \textbf{0.3158} &
  0.0789 &
  0.0526 &
  0.1579 &
  0.1196 \\
\rowcolor[HTML]{E2EFDA} 
O\&D &
  \cellcolor[HTML]{EFEFEF}0.9989 &
  0.1304 &
  0.7826 &
  0.8696 &
  \textbf{0.9130} &
  0.7826 &
  0.4783 &
  0.2174 &
  0.3913 &
  {\ul 0.9130} &
  0.3913 &
  0.8261 &
  0.6087 \\
\rowcolor[HTML]{E2EFDA} 
IM &
  \cellcolor[HTML]{EFEFEF}0.9887 &
  0.0526 &
  0.1579 &
  0.1053 &
  0.2632 &
  0.0526 &
  0.1053 &
  \textbf{0.4211} &
  0.0526 &
  0.1053 &
  0.2632 &
  0.3684 &
  0.1770 \\
\rowcolor[HTML]{E2EFDA} 
Di\&S &
  \cellcolor[HTML]{EFEFEF}0.9681 &
  0.1579 &
  0.4211 &
  0.5789 &
  0.3684 &
  0.6316 &
  0.0526 &
  0.3684 &
  0.4211 &
  0.7895 &
  {\ul 0.9474} &
  0.7368 &
  0.4976 \\
\rowcolor[HTML]{E2EFDA} 
Da\&S &
  \cellcolor[HTML]{EFEFEF}0.9394 &
  0.5263 &
  \textbf{0.5789} &
  \textbf{0.5789} &
  \textbf{0.5789} &
  0.5263 &
  0.4211 &
  0.4211 &
  \textbf{0.5789} &
  0.5263 &
  0.5263 &
  \textbf{0.5789} &
  0.5311 \\
\rowcolor[HTML]{E2EFDA} 
Rot &
  \cellcolor[HTML]{EFEFEF}0.9856 &
  0.7619 &
  0.5238 &
  {\ul 0.9048} &
  {\ul 0.9531} &
  0.8571 &
  \textbf{0.9763} &
  0.3810 &
  0.6190 &
  0.9048 &
  0.8095 &
  0.3333 &
  0.7359 \\
\rowcolor[HTML]{E2EFDA} 
Go &
  \cellcolor[HTML]{EFEFEF}0.9872 &
  0.0526 &
  0.0000 &
  0.1053 &
  0.0526 &
  0.1474 &
  \textbf{0.2000} &
  0.0000 &
  0.0000 &
  0.0684 &
  0.0263 &
  0.0000 &
  0.0593 \\
\rowcolor[HTML]{E2EFDA} 
M3G &
  \cellcolor[HTML]{EFEFEF}0.9951 &
  0.0000 &
  0.0000 &
  0.0000 &
  0.0000 &
  0.0263 &
  \textbf{0.1579} &
  0.0000 &
  0.0000 &
  0.0000 &
  0.0526 &
  0.0000 &
  0.0215 \\ \bottomrule
\end{tabular}%

\begin{tablenotes}
\footnotesize
\item \textbf{Bold values} indicate the maximum value in each row and {\ul underlined values} represent the maximum value in each
column.  
\end{tablenotes}
\end{threeparttable}
\end{table*}

\subsection{Experiment Explanations}
We use the basic prompt "Please directly solve this CAPTCHA" then upload the complete CAPTCHA images to LVLMs and let them solve the problem directly. The experimental results are shown in Table \ref{tab:basic}\footnote{The abbreviations used for sub-tasks are shown in Table \ref{full-abbr} in Supplementary Material.} .  Each row represents a sub-task, with different colors corresponding to the main-task in Figure \ref{fig:capture_details} and Figure \ref{fig:CAPTURE types}. \textbf{(Human Acc)} represents the accuracy of CAPTCHA tests completed by testers we invited. \textbf{LVLMs AVG} represents the average accuracy for each sub-task across different LVLMs. We noticed that LVLMs performed poorly in many tasks. Therefore, we further propose a two-stage framework \textbf{CRRD} in Section \ref{section:CRRD} and we attempt to improve the ability of LVLMs to solve CAPTCHAs through input and prompt-based methods.

\subsection{Enhancement Framework} \label{section:CRRD}

\begin{figure}[htbp]
    \includegraphics[width=1.05\columnwidth]{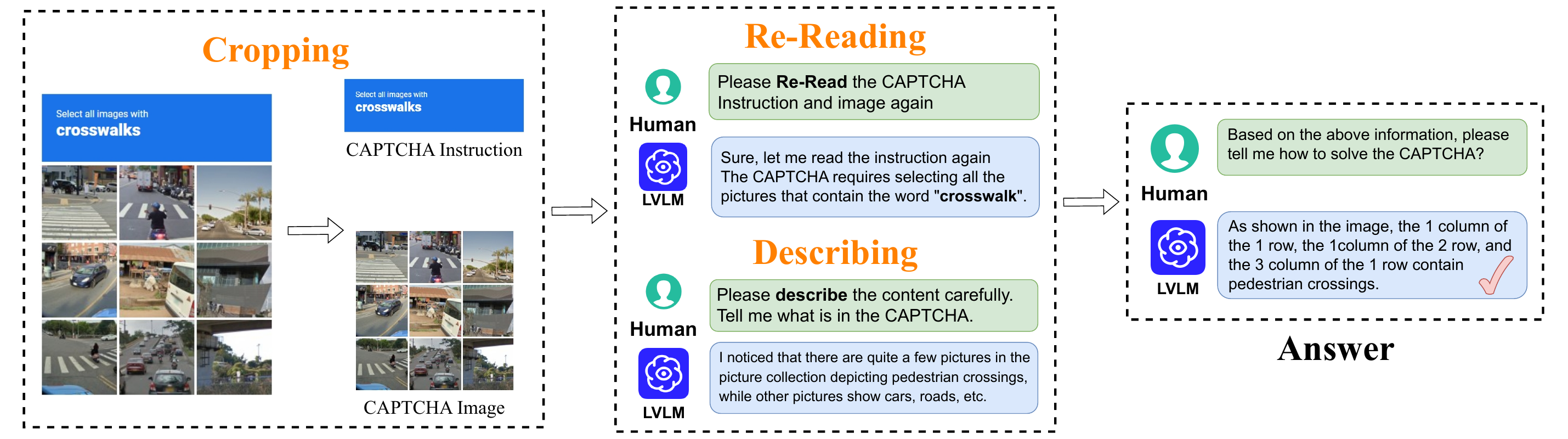}
    \caption{Overview of CRRD frame work}
    \label{fig:CRRD framework}
\end{figure}

To further explore the potential of LVLMs in solving CAPTCHAs, We attempt to make enhancement. Intuitively, LVLMs could simulate humans' visual and thinking abilities, and we propose a two-stage optimization framework named CRRD (\textbf{C}ropping, \textbf{R}e-\textbf{R}eading and \textbf{D}escribing) for LVLMs simulation. The overview of framework is shown in Figure \ref{fig:CRRD framework}.

In our experiments, we observe that some LVLMs treat CAPTCHA instructions within a CAPTCHA as part of the CAPTCHA image and attempt to process these instructions as relevant elements. We propose Cropping as the first stage. It simulates the cognitive process of humans. When humans encounter complex visual images, such as CAPTCHAs, focusing on specific sections rather than the entire image helps filter out irrelevant content. Similarly, cropping images allows LVLMs to focus on the relevant areas of the CAPTCHA. This enables the model to differentiate between the CAPTCHA instructions, which specify what needs to be solved, and the CAPTCHA image itself. By removing extraneous visual elements, the model can more effectively extract the key features required to solve the CAPTCHA, thereby improving its performance.

Next, we propose the second stage Re-Reading stage. This stage is also highly effective. In human learning and problem-solving, re-reading a text or re-examining a visual problem often leads to a deeper understanding. This process heightens attention and makes it easier to identify crucial details that may have been overlooked initially. For LVLMs, the Re-Reading mechanism serves a similar function. By processing the cropped CAPTCHA image again, the model could more effectively analyze the visual patterns, context, and semantic information within the CAPTCHA. This iterative process enhances the model's ability to recognize and interpret CAPTCHA elements, ultimately leading to more accurate solutions. Overall, through these two complementary stages, Cropping and Re-Reading, the CRRD framework equips LVLMs with improved strategies to overcome the challenges posed by CAPTCHAs. The specific prompt are provided below.

\begin{center}
\footnotesize
\label{box:prompt}
\begin{tcolorbox}[colback=blue!5!white,colframe=blue!55!black,width=0.45\textwidth,title={Prompting}]
{
{
\texttt{\# {Cropping the CAPTCHA instruction and CAPTCHA image.}} \\
\texttt{\textbf{Re-Reading Prompt:} Please read the CAPTCHA Instruction and image again: \{Input Instruction\} \{Input Image\}} \\
\texttt{{\textbf{Answer:}} Let me read the instruction again...}\\
\texttt{\textbf{Describing Prompt:} Please describe the image carefully \{Input Image\}. Tell me what is in the CAPTCHA.} \\
\texttt{{\textbf{Answer:}} Sure, I notice that there are...}\\
}
}
\end{tcolorbox}
\end{center}

\subsection{Results Analysis}
We evaluated 11 types of LVLMs: GPT-4o\cite{gpt-4o}, GPT-4o-mini\cite{GPT-4o-mini}, Claude-3-5-sonnet-20240620\cite{Claude}, Doubao-1-5-vision-pro-32k\cite{Doubao}, Gemini-1.5-pro-latest\cite{gemini1.5}, Gemini-2.0-flash\cite{gemini}, GLM-4v-plus\cite{glm}, Grok-2-vision-1212\cite{GROK-2}, Qwen2.5-VL-72B-Instruct\cite{qwen2.5}, QVQ-72B-Preview\cite{QVQ}, and DeepSeek-vl2\cite{deepseekvl2}. Table \ref{tab:basic} shows the evaluation results using the basic prompt method. And Table \ref{tab:cot} presents the evaluation results after applying our proposed CRDD framework. Overall, gemini-2.0-flash performs well across various problems, while the Chinese model doubao-1-5-vision-pro-32k performs better on  tasks with Chinese characters. Next, we will analyze the experimental results based on  RQ1 and RQ2.

\subsubsection{Response to \textbf{RQ1}}

\textbf{Could LVLMs solve all types of CAPTCHAs? } 

In Table \ref{tab:basic}, we observe that compared to Human Acc, the 11 LVLMs do not completely solve all CAPTCHA problems. The different color blocks represent different Main tasks, and below we will analyze each Main task individually.

\textbf{(1) Text Task:} LVLMs perform well, with higher accuracy for Words CAPTCHAs. These CAPTCHAs consist entirely of printed words, and despite background processing and some distortion of the words, they remain easy to recognize overall. However, the accuracy for the Letters and Digitals (L\&D) task decreases. The low accuracy observed in the validation results is due to the fact that these CAPTCHAs use a mixed form of letters and digits, with correct answers being case-sensitive and employing non-standard characters. This leads LVLMs to confuse digits and letters, as well as letters that look similar in different cases. For example, "1-i-l-I, 5-s-S, 9-g-q, 0-o-O" etc. This also affects human recognition accuracy for these CAPTCHAs. As a result, Human Acc is only 87.38\%.

\textbf{(2) Visual Task:}  we first focus on the Google reCAPTCHA 4x4 Image Segmentation (4x4 IS) task. In the experiment, we noticed that almost all LVLMs could only recognize 9 segments in the image rather than 16. We infer that this is because their training data predominantly consists of 3x3 type data, with fewer 4x4 IS training examples, resulting in very low accuracy when handling this type of task. When we treat Chinese characters as images and have LVLMs evaluate them, such as in Font Recognition (FR) and Chinese Character Selection (CCS) tasks, LVLMs show significant differences in their ability to recognize Chinese character fonts. Chinese large models, such as Doubao-1.5 and Qwen-2.5, can recognize and distinguish different Chinese fonts, such as Simsun\cite{simsun} and SimKaiti\cite{kaiti}, etc. while other LVLMs struggle to distinguish between Chinese fonts. For tasks like Jigsaw Puzzle (JP), which are inherently challenging for testers, the success rate is only 87.82\%, LVLMs not only need to identify what each piece of the puzzle represents but also must assemble the disordered pieces into a coherent whole based on image semantics, making the task quite difficult.

\textbf{(3) Behavior Task:} Here, we ask LVLMs to describe how to complete behavior-based CAPTCHAs, such as where to move the slider in a slider CAPTCHA, whether the trajectory CAPTCHA can recognize the required path to draw in the image, and where to rotate the image in a rotation CAPTCHA. LVLMs can recognize the semantic information in images and provide relatively correct descriptions. The limitations of behavior-based CAPTCHAs are discussed in the Section \ref{section:limitation}.

\textbf{(4) Game Task:} We use common games such as Gomoku and LinkLink, as well as FunCAPTCHA, to test. In the Gobang Game (Go), the board size is 5x5, with pieces in colors such as black, white, yellow, and blue. LVLMs struggle to accurately recognize the positions and colors of the pieces, resulting in poor performance. In the 3-Match Game (3MG), LVLMs cannot accurately capture the subtle differences between different patterns in the 3x3 grid, which hinders their ability to make correct selections and judgments. In FunCAPTCHA, CAPTCHAs that involve determining the orientation and direction of images, fingers, or vehicles are difficult for LVLMs. They cannot identify the direction of the indicated patterns in the same way humans can.

\begin{table*}[t]
\footnotesize
\centering
\caption{LVLMs evaluation using CRDD framework.}
\label{tab:cot}
\begin{threeparttable}
\begin{tabular}{ccccccccccccccc}
\toprule
 &
&
\multicolumn{6}{c}{\textbf{Closed Sourced LVLMs}} &
\multicolumn{5}{c}{\textbf{Open Sourced LVLMs}} &
&
\\ \cline{3-13}
\multirow{-2}{*}{\textbf{\begin{tabular}[c]{@{}c@{}}Sub Types \\ (Abbr.)\end{tabular}}} &
  \multirow{-2}{*}{\textbf{\begin{tabular}[c]{@{}c@{}}Human\\ Acc\end{tabular}}} &
  \begin{tabular}[c]{@{}c@{}}GPT\\ 4o\end{tabular} &
  \begin{tabular}[c]{@{}c@{}}GPT-4o\\ mini\end{tabular} &
  \begin{tabular}[c]{@{}c@{}}Claude\\ 3.5\end{tabular} &
  \begin{tabular}[c]{@{}c@{}}Doubao\\ 1.5\end{tabular} &
  \begin{tabular}[c]{@{}c@{}}Gemini\\ 1.5\end{tabular} &
  \begin{tabular}[c]{@{}c@{}}Gemini\\ 2.0\end{tabular} &
  \begin{tabular}[c]{@{}c@{}}GLM\\ 4v\end{tabular} &
  Grok2 &
  \begin{tabular}[c]{@{}c@{}}Qwen\\ 2.5\end{tabular} &
  QVQ &
  \begin{tabular}[c]{@{}c@{}}Deep\\ Seek\end{tabular} &
  \multirow{-2}{*}{\textbf{\begin{tabular}[c]{@{}c@{}}CRRD\\ Avg\end{tabular}}} &
  \multirow{-2}{*}{\textbf{\begin{tabular}[c]{@{}c@{}}Basic\\ Avg\end{tabular}}} \\ \midrule
\rowcolor[HTML]{FFF2CC} 
L\&D &
  \cellcolor[HTML]{EFEFEF}0.8738 &
  \textbf{0.7143} &
  \textbf{0.6190} &
  0.5238 &
  0.3333 &
  \textbf{0.6190} &
  \textbf{0.6190} &
  0.5714 &
  0.6190 &
  0.6667 &
  0.5238 &
  0.4762 &
  0.5714 (+0.0043) &
  0.5671 \\
\rowcolor[HTML]{FFF2CC} 
Words &
  \cellcolor[HTML]{EFEFEF}0.9980 &
  {\ul \textbf{0.9524}} &
  0.9048 &
  0.7143 &
  0.9048 &
  \textbf{0.9524} &
  \textbf{0.9524} &
  0.8095 &
  {\ul 0.9048} &
  0.9048 &
  0.8095 &
  {\ul 0.9048} &
\textbf{{\ul 0.8831} (+0.0433)} &
  {\ul \textbf{0.8398}} \\
\rowcolor[HTML]{E1D5E7} 
3x3 EC &
  \cellcolor[HTML]{EFEFEF}0.9415 &
  \textbf{0.3158} &
  0.0526 &
  0.2105 &
  0.1842 &
  0.1053 &
  0.1579 &
  0.1842 &
  0.2105 &
  0.2895 &
  0.2105 &
  0.0789 &
  0.1818 (+0.1052) &
  0.0766 \\
\rowcolor[HTML]{E1D5E7} 
3x3 IC &
  \cellcolor[HTML]{EFEFEF}0.9363 &
  0.6842 &
  0.2895 &
  0.4211 &
  0.6842 &
  0.7632 &
  {\ul \textbf{0.9149}} &
  0.3947 &
  0.7632 &
  0.5000 &
  0.4737 &
  0.2105 &
  0.5646 (+0.0718) &
  0.4928 \\
\rowcolor[HTML]{E1D5E7} 
4x4 IS &
  \cellcolor[HTML]{EFEFEF}0.9650 &
  0.3421 &
  0.3684 &
  0.3421 &
  \textbf{0.3421} &
  0.3947 &
  0.3684 &
  0.3421 &
  0.3158 &
  \textbf{0.3421} &
  0.2895 &
  0.3421 &
  0.3445 (+0.1005) &
  0.2440 \\
\rowcolor[HTML]{E1D5E7} 
CCI &
  \cellcolor[HTML]{EFEFEF}0.9952 &
  0.5263 &
  0.5263 &
  0.5826 &
  0.5662 &
  \textbf{0.7895} &
  0.7368 &
  0.5263 &
  0.6316 &
  0.1579 &
  0.3684 &
  0.5789 &
  0.5446 (+0.0569) &
  0.4877 \\
\rowcolor[HTML]{E1D5E7} 
LI &
  \cellcolor[HTML]{EFEFEF}0.9993 &
  0.0119 &
  \textbf{0.0168} &
  0.0159 &
  0.0020 &
  0.0089 &
  0.0119 &
  0.0000 &
  0.0119 &
  0.0159 &
  0.0000 &
  0.0119 &
  0.0097 (+0.0009) &
  0.0088 \\
\rowcolor[HTML]{E1D5E7} 
IP &
  \cellcolor[HTML]{EFEFEF}0.9761 &
  0.5000 &
  0.3947 &
  0.4211 &
  \textbf{0.8684} &
  \textbf{0.9474} &
  \textbf{0.6842} &
  0.5789 &
  0.0263 &
  0.2895 &
  0.1579 &
  0.0789 &
  0.4498 (+0.0718) &
  0.3780 \\
\rowcolor[HTML]{E1D5E7} 
DS &
  \cellcolor[HTML]{EFEFEF}0.9601 &
  0.7895 &
  0.6842 &
  0.9474 &
  \textbf{0.9497} &
  0.8947 &
  0.9474 &
  0.7895 &
  0.2632 &
  0.3158 &
  0.4737 &
  0.4211 &
  0.6890 (+0.1340) &
  0.5550 \\
\rowcolor[HTML]{E1D5E7} 
SR &
  \cellcolor[HTML]{EFEFEF}0.9130 &
  {\ul \textbf{0.9524}} &
  0.8571 &
  \textbf{0.9524} &
  \textbf{0.9524} &
  0.8571 &
  0.8095 &
  {\ul \textbf{0.9524}} &
  0.5238 &
  0.9048 &
  {\ul 0.9048} &
  {\ul \textbf{0.9524}} &
  0.8745 (+0.1256) &
  0.7489 \\
\rowcolor[HTML]{E1D5E7} 
JP &
  \cellcolor[HTML]{EFEFEF}0.8782 &
  0.1842 &
  0.0789 &
  0.0789 &
  0.1053 &
  \textbf{0.2368} &
  \textbf{0.1842} &
  0.1316 &
  0.1316 &
  0.1842 &
  0.0526 &
  0.0000 &
  0.1244 (+0.1148) &
  0.0096 \\
\rowcolor[HTML]{E1D5E7} 
BAS &
  \cellcolor[HTML]{EFEFEF}0.9849 &
  0.1316 &
  0.1316 &
  \textbf{0.2368} &
  \textbf{0.1316} &
  0.1842 &
  0.1579 &
  0.0789 &
  0.1842 &
  \textbf{0.2368} &
  0.0789 &
  0.1053 &
  0.1507 (+0.0933) &
  0.0574 \\
\rowcolor[HTML]{E1D5E7} 
IS &
  \cellcolor[HTML]{EFEFEF}0.9582 &
  0.5526 &
  0.2895 &
  0.5526 &
  0.7368 &
  0.7895 &
  \textbf{0.8684} &
  0.3158 &
  0.2105 &
  0.2895 &
  0.0789 &
  0.2632 &
  0.4498 (+0.2345) &
  0.2153 \\
\rowcolor[HTML]{E1D5E7} 
CCS &
  \cellcolor[HTML]{EFEFEF}0.9613 &
  0.4211 &
  0.0000 &
  0.2895 &
  0.2895 &
  0.2895 &
  \textbf{0.4474} &
  0.2632 &
  0.0000 &
  0.3158 &
  0.1053 &
  0.1842 &
  0.2368 (+0.2009) &
  0.0359 \\
\rowcolor[HTML]{E1D5E7} 
FR &
  \cellcolor[HTML]{EFEFEF}0.9890 &
  0.4474 &
  0.2368 &
  0.2895 &
  0.5000 &
  0.7368 &
  \textbf{0.7895} &
  0.3421 &
  0.1316 &
  0.2895 &
  0.0789 &
  0.3684 &
  0.3828 (+0.0646) &
  0.3182 \\
\rowcolor[HTML]{FFCCC9} 
\cellcolor[HTML]{F8CECC}Slide &
  \cellcolor[HTML]{EFEFEF}0.9973 &
  0.8684 &
  {\ul \textbf{0.9688}} &
  0.8684 &
  0.7895 &
  0.9211 &
  \textbf{0.8947} &
  0.7368 &
  0.7895 &
  0.8158 &
  0.6053 &
  0.5789 &
  0.8034 (\textbf{+0.6957}) &
  0.1077 \\
\rowcolor[HTML]{FFCCC9} 
\cellcolor[HTML]{F8CECC}Tra &
  \cellcolor[HTML]{EFEFEF}0.9733 &
  \textbf{0.7368} &
  0.6579 &
  0.6053 &
  0.5526 &
  0.7368 &
  \textbf{0.6579} &
  0.4211 &
  0.4474 &
  0.3947 &
  0.6842 &
  0.4474 &
  0.5766 (+0.1340) &
  0.4426 \\
\rowcolor[HTML]{FFCCC9} 
\cellcolor[HTML]{F8CECC}IRR &
  \cellcolor[HTML]{EFEFEF}0.9570 &
  0.5789 &
  0.5789 &
  0.4737 &
  0.5789 &
  0.4474 &
  \textbf{0.6579} &
  0.0526 &
  \textbf{0.6579} &
  0.5789 &
  0.1316 &
  0.2895 &
  0.4569 (+0.3373) &
  0.1196 \\
\rowcolor[HTML]{E2EFDA} 
O\&D &
  \cellcolor[HTML]{EFEFEF}0.9989 &
  0.7391 &
  0.7826 &
  \textbf{0.9130} &
  \textbf{0.3043} &
  0.7391 &
  0.7826 &
  0.7391 &
  0.8696 &
  {\ul 0.6957} &
  0.8696 &
  0.8261 &
  0.7510 (+0.1423) &
  0.6087 \\
\rowcolor[HTML]{E2EFDA} 
IM &
  \cellcolor[HTML]{EFEFEF}0.9887 &
  \textbf{0.4211} &
  0.3684 &
  \textbf{0.4211} &
  0.4211 &
  0.3158 &
  0.3158 &
  \textbf{0.4211} &
  0.4211 &
  0.2632 &
  0.3684 &
  0.3158 &
  0.3684 (+0.1914) &
  0.1770 \\
\rowcolor[HTML]{E2EFDA} 
Di\&S &
  \cellcolor[HTML]{EFEFEF}0.9681 &
  0.8947 &
  0.8947 &
  0.7368 &
  0.9474 &
  0.6316 &
  0.7895 &
  0.9474 &
  0.7895 &
  0.8421 &
  {\ul 0.8947} &
  0.6316 &
  0.8182 (+0.3206) &
  0.4976 \\
\rowcolor[HTML]{E2EFDA} 
Da\&S &
  \cellcolor[HTML]{EFEFEF}0.9394 &
  0.5263 &
  \textbf{0.7368} &
  \textbf{0.6316} &
  \textbf{0.4737} &
  0.3158 &
  0.4737 &
  0.3684 &
  \textbf{0.4737} &
  0.6316 &
  0.4737 &
  \textbf{0.5263} &
  0.5120 (-0.0191) &
  0.5311 \\
\rowcolor[HTML]{E2EFDA} 
Rot &
  \cellcolor[HTML]{EFEFEF}0.9856 &
  0.7143 &
  0.6190 &
  {\ul 0.9579} &
  {\ul 0.9562} &
  {\ul 0.9524} &
  \textbf{0.9048} &
  0.6190 &
  {\ul 0.9048} &
  0.9596 &
  0.8571 &
  0.5238 &
  0.8190 (+0.0831) &
  0.7359 \\
\rowcolor[HTML]{E2EFDA} 
Go &
  \cellcolor[HTML]{EFEFEF}0.9872 &
  0.1211 &
  0.0526 &
  0.1579 &
  0.0789 &
  0.1842 &
  \textbf{0.1579} &
  0.0789 &
  0.0947 &
  0.0526 &
  0.0000 &
  0.1053 &
  0.0986 (+0.0393) &
  0.0593 \\
\rowcolor[HTML]{E2EFDA} 
M3G &
  \cellcolor[HTML]{EFEFEF}0.9951 &
  0.0526 &
  0.0526 &
  0.0789 &
  0.1053 &
  0.1579 &
  \textbf{0.0789} &
  0.0526 &
  0.1579 &
  0.0526 &
  0.1053 &
  0.0263 &
  0.0837 (+0.0622) &
  0.0215 \\ \bottomrule
\end{tabular}%

\begin{tablenotes}
\footnotesize
\item \textbf{Bold values} indicate the maximum value in each row and {\ul underlined values} represent the maximum value in each
column.  
\end{tablenotes}
\end{threeparttable}
\end{table*}

\subsubsection{Response to \textbf{RQ2}}
\textbf{RQ2: How about input methods and prompt  techniques enhance LVLMs’ ability to solve CAPTCHAs? }

For RQ2, we proposed the CRRD framework to enhance the reasoning ability of LVLMs, thereby improving their ability to solve CAPTCHAs. The experimental results are shown in Table \ref{tab:cot}. Basic Avg corresponds to the LVLMs Avg column in Table \ref{tab:basic}. First, comparing the average accuracy, the CRRD Avg value in Table \ref{tab:cot} shows an improvement over Basic Avg. Among these, the sliding CAPTCHA (Slide) in the behavior-based task showed the greatest improvement of 69.57\%, while other tasks also saw some improvement. The accuracy for individual models on specific tasks in Table \ref{tab:cot} has improved significantly compared to Table \ref{tab:basic}. This indicates that the CRRD framework we proposed yields the best results, improving the accuracy of LVLMs in solving CAPTCHAs. However, compared to Human Acc, the accuracy still does not reach human-level performance in solving CAPTCHAs. This suggests that despite using the proposed CRRD framework, LVLMs still lack the ability to solve CAPTCHAs at the human level.

\section{Limitation and Future Work}\label{section:limitation}


Our study, which focuses on leveraging LVLMs for CAPTCHA resolution, is subject to an inherent limitation. Currently, LVLMs lack the native ability to physically interact with CAPTCHA elements, such as in the case of slider and rotation CAPTCHAs. These types of CAPTCHAs require an active manipulation of visual components, which is beyond the scope of what LVLMs can directly execute. To overcome this issue, we devise an alternative approach for evaluating LVLMs' understanding of the behavior CAPTCHAs. Instead of direct interaction, we analyze the LVLMs' outputs. By examining the semantic and visual-reasoning based descriptions and solutions proposed by the LVLMs, we inferre their comprehension of the CAPTCHA content and their ability to formulate a proper resolution strategy. For example, when dealing with a rotation CAPTCHA, if an LVLM can accurately describe the required rotation angle and the object to be rotated in its output, it indicates a certain level of understanding of the problem. This method, although indirect, still provides valuable insights. It aligns with the fundamental goal of assessing the cognitive-like capabilities of LVLMs in handling complex tasks related to CAPTCHAs. By focusing on the models' understanding and cognitive processing rather than physical interaction, we can still gauge their progress in emulating human like visual and reasoning skills. Moreover, the analysis of textual outputs allows for a standardized and replicable evaluation across different LVLMs, which is crucial for a comprehensive benchmarking study.

Currently, researchers are already using Function Call (FC)\footnote{https://platform.openai.com/docs/guides/function-calling} and Model Context Protocol (MCP)\footnote{https://modelcontextprotocol.io/introduction} to expand the capabilities of LVLMs, allowing them to interact with external systems. In the future, LVLMs will utilize FC and MCP to perform the physical operations on CAPTCHAs. Hence, the emegence of our \(\textit{CAPTURE}\) benchmark lays a solid foundation for future research in this area, facilitating the development of such enhanced LVLMs with CAPTCHA - manipulation capabilities.

\section{Conclusion}

With rapid advancements, LVLMs have developed the ability to simulate human-like visual and reasoning capabilities, including solving CAPTCHAs, a widely used method for distinguishing between automated bots and humans. As CAPTCHAs evolve, the ability of LVLMs to solve various types remains a challenge. Thus, We propose \textit{CAPTURE}, a comprehensive benchmark for evaluating LVLMs' ability to solve CAPTCHAs. In this study, we use the \textit{CAPTURE} benchmark, which includes 4 main CAPTCHA types and 25 sub-types from 31 vendors, to evaluate 11 LVLMs. We also propose a two-stage framework, CRRD, to improve the accuracy of LVLMs in CAPTCHA-solving tasks. The framework and methods we propose could improve accuracy by up to 69\%. However, LVLMs still show lower accuracy than humans in solving CAPTCHAs. This indicates existing LVLMs are still unable to fully simulate human visual and reasoning abilities to solve all current CAPTCHAs, due to the inherent characteristics of LVLM models.

\bibliographystyle{ACM-Reference-Format}
\bibliography{main}

\clearpage
\appendix
\begin{table*}[h!]
\caption{The details of previous CAPTCHA-related studies and datasets and our work (from 2009 to 2025).}
\label{tab:vendors}
\resizebox{\textwidth}{!}{%
\begin{tabular}{lccccccclcc}
\hline
\multicolumn{1}{c}{\multirow{2}{*}{Dataset}} &
  \multicolumn{7}{c}{CAPTCHA Types} &
  \multicolumn{1}{c}{\multirow{2}{*}{CAPTCHA Source}} &
  \multicolumn{1}{c}{\multirow{2}{*}{CAPTCHAs}} &
  \multirow{2}{*}{Year} \\  \cline{2-8}
\multicolumn{1}{c}{} &
  Text &
  Images &
  Games &
  Behavior &
  Audio &
  Video &
  Others &
  \multicolumn{1}{c}{} &
   &
   \\ \hline
Bigham et.al \cite{Bigham} &
  \checkmark &
   &
   &
   &
  \checkmark &
   &
  Interface &
  Alexa \cite{Alexa} &
  2,350 &
  2009 \\
Bursztein et.al \cite{Bursztein} &
  \checkmark &
   &
   &
   &
  \checkmark &
   &
   &
  Newly proposed &
  318,000 &
  2010 \\
Ross et.al \cite{Ross} &
   &
  \checkmark &
   &
   &
   &
   &
  Sketch (2D/3D) &
  Newly proposed &
  14,032 &
  2010 \\
Gao et.al \cite{JigsawPuzzle} &
   &
   &
  \checkmark &
   &
   &
   &
   &
  Newly proposed &
  300 &
  2010 \\
Ho et.al \cite{Ho}&
  \checkmark &
   &
   &
   &
   &
   &
   &
  Major websites &
  1,407,500 &
  2011 \\
Fidas et.al \cite{Fidas} &
  \checkmark &
   &
   &
   &
   &
   &
   &
  Newly proposed &
  210 &
  2011 \\
Mohamed et.al \cite{Mohamed} &
   &
   &
  \checkmark &
   &
   &
   &
   &
  Newly proposed &
  480 &
  2014 \\
Krol et.al \cite{Krol} &
  \checkmark &
   &
  \checkmark &
   &
   &
   &
   &
  \cite{reCAPTCHAv2}, \cite{PlayThru} &
  261 &
  2016 \\
Jain et.al \cite{Jain} &
   &
   &
   &
   &
  \checkmark &
   &
   &
  Newly proposed &
  4,740 &
  2019 \\
Gao et.al \cite{Gao}&
  \checkmark &
  \checkmark &
   &
   &
   &
   &
   &
  \cite{Camouflageimages}, \cite{NuCaptcha}, \cite{xu2012security}, \cite{Motion-Based-CAPTCHAs} &
  3,600 &
  2019 \\
Tanthavech et.al \cite{Tanthavech} &
  \checkmark &
  \checkmark &
  \checkmark &
  \checkmark &
   &
   &
  \multicolumn{1}{l}{} &
  Websites &
  200 &
  2019 \\
Feng et.al \cite{Feng} &
  \checkmark &
  \checkmark &
   &
   &
  \checkmark &
  \checkmark &
  Mobile Sensor &
  Newly proposed, \cite{dapphp}. &
  4,920 &
  2020 \\
Gao et.al \cite{Visual-Reasoning-CAPTCHA} &
   &
  \checkmark &
   &
   &
   &
   &
   &
  Major websites &
  13,500 &
  2021 \\
Searles et.al \cite{Searles} &
  \checkmark &
  \checkmark &
  \checkmark &
  \checkmark &
   &
   &
   &
  Alexa \cite{Alexa} &
  14,000 &
  2023 \\
Xu et.al \cite{ViRC} &
   &
  \checkmark &
   &
   &
   &
   &
   &
  Major websites &
  40,300 &
  2023 \\
Deng et.al \cite{Oedipus} &
   &
  \checkmark &
  \checkmark &
   &
   &
   &
   &
  Major websites &
  50 &
  2024 \\
Ding et.al \cite{IllusionCAPTCHA} &
  \checkmark &
  \checkmark &
  \checkmark &
   &
   &
   &
   &
  Major websites &
  330 &
  2025 \\
\textbf{Ours} &
  \checkmark &
  \checkmark &
  \checkmark &
  \checkmark &
    &
    &
    &
  \textbf{31 CAPTCHA vendors} &
  \textbf{61,464} &
  \textbf{2025} \\ \hline
\end{tabular}%
}\end{table*}

\begin{table*}[h!]
\caption{31 Vendors list.}
\label{tab:vendors}
\centering
\begin{tabular}{ll}
\toprule
\multicolumn{1}{c}{\textbf{International}} & \textbf{China}  \\ \midrule
Cloudflare Turnstile     & Dingxiang     \\
Google reCAPTCHA         & DeepCleer     \\
hCaptcha                 & NetEase Yidu  \\
Arkose Labs FunCAPTCHA   & Geetest       \\
DataDome CAPTCHA         & Aliyun        \\
PerimeterX               & Tencent T-Sec \\
AWS (Amazon) WAF Captcha & Xiaohongsh    \\
MTCaptcha                & Xiaodun       \\
Lemin CAPTCHA            & Yunpian       \\
Yandex SmartCaptcha      & 360 Tianyu    \\
mCaptcha                 & Luosimao      \\
KeyCAPTCHA               & Vaptcha       \\
CyberSiARA               & V5            \\
Friendly Captcha         & Kyger         \\
ARCaptcha                & FastYotest    \\
CaptchaFox               &               \\ \bottomrule
\end{tabular}%
\end{table*}
\clearpage

\end{sloppypar}
\end{document}